\documentclass[10pt,twocolumn,letterpaper]{article}

\usepackage[accsupp]{axessibility}  

\usepackage{iccv}
\usepackage{times}
\usepackage[utf8]{inputenc} 
\usepackage[T1]{fontenc}    
\usepackage{url}            
\usepackage{booktabs}       
\usepackage{amsfonts}       
\usepackage{nicefrac}       
\usepackage{microtype}      
\usepackage{xcolor}         

\usepackage{graphicx}
\usepackage{amsmath,amssymb}
\usepackage{verbatim}
\usepackage{multirow}
\usepackage{marvosym}
\usepackage{makecell}
\usepackage{bm}
\usepackage{wrapfig}
\usepackage{caption}
\usepackage{subfigure}

\usepackage[pagebackref=true,breaklinks=true,letterpaper=true,colorlinks,bookmarks=false]{hyperref}

\iccvfinalcopy 

\def\iccvPaperID{4615} 

\ificcvfinal\fi

\begin{document}

\title{Auxiliary Tasks Benefit 3D Skeleton-based Human Motion Prediction}

\author{Chenxin Xu\textsuperscript{1,2}, Robby T. Tan\textsuperscript{2}, Yuhong Tan\textsuperscript{1}, Siheng Chen\textsuperscript{1,3\footnotemark[1]},  Xinchao Wang\textsuperscript{2}, Yanfeng Wang\textsuperscript{3,1}
\\\textsuperscript{1}Shanghai Jiao Tong University,  
\textsuperscript{2}National University of Singapore,
\textsuperscript{3}Shanghai AI Laboratory
\\
{\tt\small {\{xcxwakaka,tyheeeer,sihengc,wangyanfeng\}@sjtu.edu.cn},} 
\tt\small {\{robby.tan,xinchao\}@nus.edu.sg}
}

\maketitle
\renewcommand{\thefootnote}{\fnsymbol{footnote}}
\footnotetext[1]{Corresponding author.}

\begin{abstract}
Exploring spatial-temporal dependencies from observed motions is one of the core challenges of human motion prediction. Previous methods mainly focus on dedicated network structures to model the spatial and temporal dependencies. This paper considers a new direction by introducing a model learning framework with auxiliary tasks. 
In our auxiliary tasks, partial body joints' coordinates are corrupted by either masking or adding noise and the goal is to recover corrupted coordinates depending on the rest coordinates. To work with auxiliary tasks, we propose a novel auxiliary-adapted transformer, which can handle incomplete, corrupted motion data and achieve coordinate recovery via capturing spatial-temporal dependencies. Through auxiliary tasks, the auxiliary-adapted transformer is promoted to capture more comprehensive spatial-temporal dependencies among body joints' coordinates, leading to better feature learning. Extensive experimental results have shown that our method outperforms state-of-the-art methods by remarkable margins of 7.2\%, 3.7\%, and 9.4\% in terms of 3D mean per joint position error (MPJPE) on the Human3.6M, CMU Mocap, and 3DPW datasets, respectively. We also demonstrate that our method is more robust under data missing cases and noisy data cases. Code is available at \url{https://github.com/MediaBrain-SJTU/AuxFormer}.


\end{abstract}

\section{Introduction}


3D skeleton-based human motion prediction aims to forecast future human motions based on past observations, which has a wide range of applications, such as human-machine interaction \cite{gui2018teaching,gorecky2014human,xu2021invariant} and autonomous driving \cite{levinson2011towards,chen20203d,liang2020learning}. One of the main challenges of this problem is extracting spatial-temporal dependencies among observed motions to make feature representations more informative. These dependencies arise due to the complex interactions between different joints and the temporal dynamics of motion. Therefore, developing effective methods to capture these dependencies is crucial for a more accurate 
human motion prediction.

\begin{figure}[t] 
\centering
\includegraphics[width=0.47\textwidth]{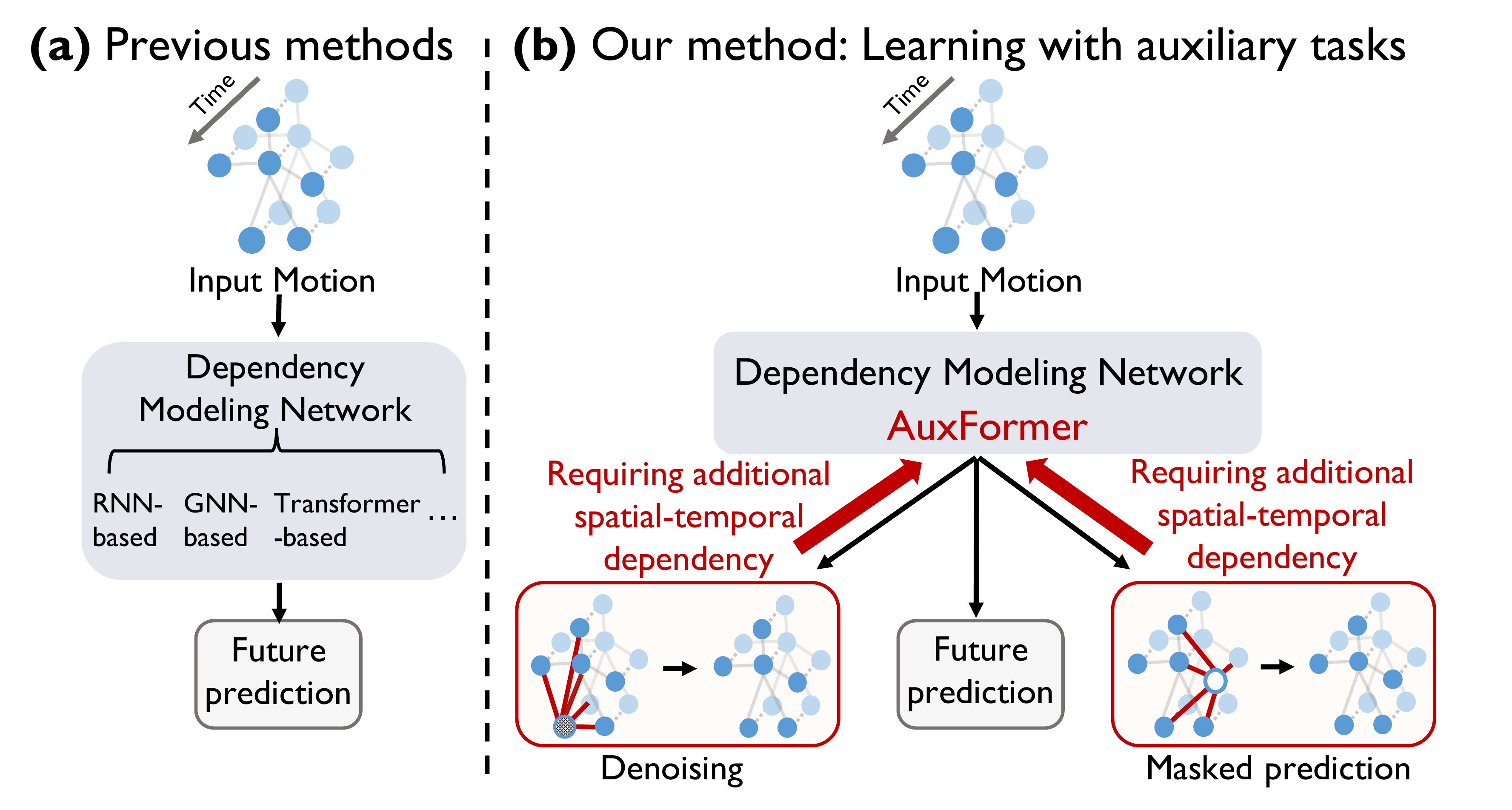}
\vspace{-3mm}
\caption{\small Compared to previous methods with various operation designs, we propose a new direction: an auxiliary learning framework that incorporates auxiliary tasks, including denoising and masking prediction. These auxiliary tasks impose additional requirements through recovery,  forcing the network to exploit spatial-temporal dependencies more comprehensively.}
\label{fig:jewel}
\vspace{-5mm}
\end{figure}

Existing methods proposed dedicated network structures to model the spatial and temporal dependencies. A few methods use RNN \cite{martinez2017human,gui2018adversarial} and TCN \cite{cui2020learning} structures to model temporal dependencies, but neglect the spatial ones. 
To learn spatial dependencies between joints, \cite{mao2019learning} proposes a GCN network with learnable weights where nodes are body joints. Following the GCN design, DMGNN \cite{li2020dynamic} and MSR-GCN \cite{dang2021msr} further build multiscale body graphs to model local-global spatial features. PGBIG \cite{ma2022progressively} additionally proposes temporal graph convolutions to extract spatial-temporal
features. SPGSN \cite{li2022skeleton} proposes a graph scattering network to further model temporal dependencies from multiple graph spectrum bands. Besides GCN-based methods, transformer architectures \cite{aksan2021spatio,cai2020learning} are also used to model pair-wise spatial-temporal dependencies. The structures of models to capture spatial-temporal dependencies have been extensively studied. This raises a natural question: can we enhance spatial-temporal dependency modeling from other perspectives?

In this paper, besides network structures, we further consider an orthogonal approach: proposing a new auxiliary model learning framework by adding auxiliary tasks to promote better learning of spatial-temporal dependencies. 
The proposed framework jointly learns the primary motion prediction task along with additional auxiliary tasks, with sharing the same dependency modeling network. The core idea of the auxiliary tasks is to corrupt partial observed coordinates and set a goal to recover corrupted coordinates using correlated normal coordinates according to their spatial-temporal dependencies.
The goals of the auxiliary tasks are highly correlated with the primary prediction task, as they both require the network to model spatial-temporal dependencies effectively. 
Therefore, through the additional requirements imposed by the auxiliary tasks, the dependency modeling network is forced to learn more effective and comprehensive spatial-temporal dependencies.  Our learning framework complements existing methods by further emphasizing the effective learning of the network structure.




To be specific, we introduce two kinds of auxiliary tasks: a denoising task and a masked feature prediction task. The denoising task randomly adds noise into joint coordinates at different timestamps in the input motion, and the aim is to recover the original input motion. The masked feature prediction task randomly masks joint coordinates at different timestamps, and the goal is to predict the masked joint positions. 
Comparing to previous popular methods based on masking/denoising autoencoding like masked autoencoder (MAE) \cite{he2022masked} and denoising autoencoder (DAE) \cite{vincent2008extracting}, which are mainly designed for model pre-training, we treat the denoising and masking prediction as auxiliary tasks to aid the primary fully-supervised motion prediction task and jointly learn all the tasks together. Moreover, previous methods using masking/denoising strategies mostly focus on data types of images \cite{he2022masked,mahajan2018exploring}, videos \cite{feichtenhofer2022masked,tan2021vimpac}, languages \cite{devlin2018bert,radford2018improving,TaskRes} and point clouds \cite{pang2022masked,wang2021pwclo}, but rarely on motion sequences, especially human motions. Our work fills this gap and utilizes the strategies to promote more effective spatial-temporal dependencies learning in human motion prediction. 

To cooperate with auxiliary tasks in the learning framework, the dependency modeling network structure faces two demands. First, the network has to learn spatial-temporal dependencies between the corrupted coordinate and the normal coordinate on a coordinate-wise basis to enable recovery. Second, the network has to be adaptive to incomplete motions, caused by the masking prediction task. 
Thus, we specifically propose an auxiliary-adapted transformer network to meet both two demands. 
To model the coordinate-wise dependency, we consider each coordinate as one individual feature in our network and use spatial-temporal attention that models spatial-temporal dependencies between two coordinates' features. To be adaptive to incomplete data, we add tokens into the masked coordinates' feature to inform the network that the data is missing, and incorporate a mask-aware design into spatial-temporal attention that enables arbitrary incomplete inputs. Integrating the above learning framework and network design, we name our method \textit{AuxFormer}.

We conduct experiments on both short-term and long-term motion prediction on large-scale datasets: Human3.6M \cite{ionescu2013human3}, CMUMocap and 3DPW \cite{von2018recovering}.
Our method significantly outperforms state-of-the-art methods in terms of mean per joint position error (MPJPE). We also show our method is more robust under data missing and noisy cases. 
The main contributions of our work are summarized here:

$\bullet$ We propose a new auxiliary learning framework for human motion prediction to jointly learn the prediction task with two auxiliary tasks: denoising and masking prediction.
Through auxiliary tasks, the model network is forced to learn more comprehensive spatial-temporal dependencies.

$\bullet$ We propose an auxiliary-adapted transformer to cooperate with the learning framework. The auxiliary-adapted transformer models coordinate-wise spatial-temporal dependencies and is adaptive to incomplete motion data.

$\bullet$ We conduct experiments to verify that our method significantly outperforms existing works by 7.2\%/3.7\%/9.4\%. We also show our method is more robust under data missing cases and noisy data cases.

\begin{figure*}[t] 
\centering
\includegraphics[width=0.96\textwidth]{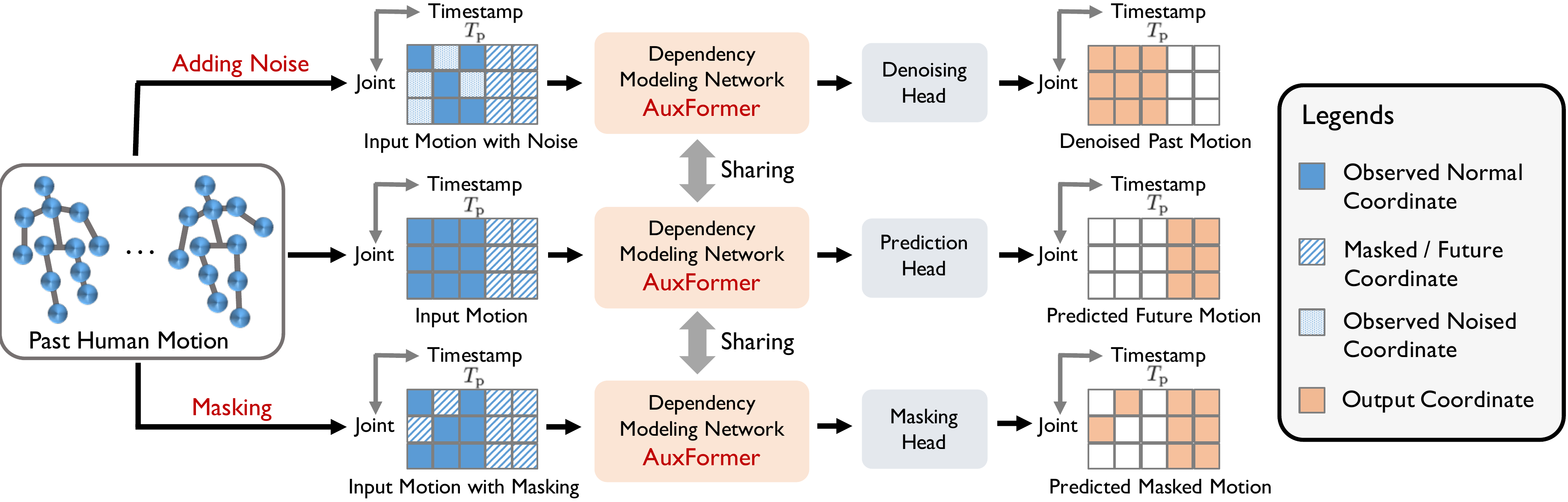}
\vspace{-3.5mm}
\caption{\small The learning framework architecture. The framework consists of three tasks: the primary future prediction task (middle branch), the auxiliary denoising task (upper branch), and the auxiliary masking prediction task (lower branch). These three tasks share the same dependency modeling network and use different heads.}
\label{fig:framework}
\vspace{-4mm}
\end{figure*}

\section{Related Work}
\noindent\textbf{Human Motion Prediction.}
Early methods for human motion prediction captured shallow temporal dynamics using state prediction \cite{lehrmann2014efficient,taylor2009factored}. Later, RNN-based models like ERD \cite{fragkiadaki2015recurrent}, Pose-VAE \cite{walker2017pose}, Structural-RNN \cite{jain2016structural}, \cite{gui2018adversarial}, Res-sup \cite{martinez2017human} and AHMR \cite{liu2022investigating} improved temporal dependency modeling.
\cite{guo2019human,li2018convolutional} use spatial convolutions and predict whole sequences without accumulation. 
More recent approaches use graph convolutional networks \cite{kipf2016semi,chen2020simple,xu2022groupnet,yang2020CVPR,ZhangyangGraph1,ZhangyangGraph2} to represent human poses as graphs and capture spatial dependencies. Researchers build fully-connected graphs \cite{mao2019learning,cui2020learning,salzmann2022motron}, multi-scale graphs \cite{li2020dynamic,dang2021msr}, and semi-constrained graphs \cite{liu2021motion} to encode skeletal connections and prior knowledge. \cite{sofianos2021space,ma2022progressively} extend graph convolution to temporal dimension to model both temporal and spatial dependencies. \cite{li2022skeleton} proposes graph scattering networks to decompose pose features into richer graph spectrum bands. \cite{cui2021towards} proposes a multi-task graph convolutional network for the incomplete observations cases. EqMotion \cite{xu2023eqmotion} proposes an equivariant motion prediction model to achieve motion equivariance and interaction invariance based on graphs. Besides GCN-based methods, transformer structures \cite{aksan2021spatio,cai2020learning} also are adapted to model pairwise spatial-temporal dependencies. Unlike most previous methods that focus on dedicated network structures, we adopt a novel strategy by introducing an auxiliary learning framework to efficiently train a network for better performance.

\vspace{0.1cm}
\noindent\textbf{Denoising and Masked Autoencoding.}
Both denoising and masked autoencoding aim to learn representative features by reconstructing corrupted data. Denoising autoencoding, first proposed by \cite{vincent2008extracting}, makes learned representations robust to partial corruption. Different corruptions are further proposed like gray colorization \cite{zhang2016colorful}, pixel masking \cite{mahajan2018exploring}, and channel splitting \cite{zhang2017split}. Masked autoencoding applies region masking as a corruption and is a special form of denoising autoencoding. Motivated by large language pre-training models like BERT \cite{devlin2018bert} and GPT \cite{radford2018improving,radford2019language}, masked image prediction methods are proposed for vision model pre-training with different reconstruction targets like image pixels \cite{chen2020generative,dosovitskiy2020image,xie2022simmim,bachmann2022multimae,DBLP:conf/nips/Liu0YYW22}, discrete tokens \cite{bao2021beit,zhou2021ibot,Xinyin2023structural}, and deep features \cite{baevski2022data2vec,wei2022masked}. \cite{he2022masked} presents the masked autoencoder (MAE) to accelerate model pre-training by masking a high proportion of input images and developing an asymmetric encoder-decoder architecture. This idea is used in other data types such as point clouds \cite{pang2022masked,zanjani2019mask,wang2021pwclo,yu2022point} and videos \cite{feichtenhofer2022masked,tan2021vimpac,gabeur2022masking}. Our method focuses on motion sequence data and incorporates masking/denoising as auxiliary tasks to assist the motion prediction task, different from previous pre-training methods.

\noindent\textbf{Multi-Task Learning and Auxiliary Learning.}
Multi-task learning \cite{zhang2021survey,caruana1998multitask} aims to improve tasks' performance by sharing information between multiple relevant tasks. Similarly, auxiliary learning uses multiple auxiliary tasks to assist the primary task, but it only focuses on improving the performance of the primary task. Auxiliary learning has been used in various tasks like semantic segmentation \cite{xu2021leveraging}, scene understanding \cite{liebel2018auxiliary}, image and video captioning \cite{hosseinzadeh2021image,gao2021hierarchical}, speech recognition \cite{toshniwal2017multitask}, view synthesis \cite{flynn2016deepstereo}, and reinforcement learning \cite{jaderberg2016reinforcement} based on images, videos, and languages. However, auxiliary learning has not been extensively applied in 3D human motion prediction. To the best of our knowledge, this paper is the first to introduce auxiliary learning to 3D human motion prediction.



\section{Prediction Framework with Auxiliary Tasks}
\subsection{Problem Formulation}
The 3D skeleton-based human motion task aims to predict future human motions given past motions. Mathematically, let 
$\mathbf{X}^t = [\mathbf{x}_1^t,\mathbf{x}_2^t,\cdots \mathbf{x}_J^t]\in \mathbb{R}^{J\times 3}$ denotes the human pose consisting 3D coordinates of $M$ body joints at timestamp $t$, where $\mathbf{x}_j^t \in \mathbb{R}^3$ is the $j$th joint's coordinate. Let $\mathbb{X}^{-} = [\mathbf{X}^{1},\mathbf{X}^{2},\cdots,\mathbf{X}^{T_{\rm p}}] \in \mathbb{R}^{T_{\rm p}\times J \times 3}$  be the past motion and $\mathbb{X}^{+} = [\mathbf{X}^{T_{\rm p}+1},\mathbf{X}^{T_{\rm p}+2},\cdots,\mathbf{X}^{T_{\rm p}+T_{\rm f}}] \in \mathbb{R}^{T_{\rm f}\times J \times 3}$ be the future motion. $T_{\rm p}$ and $T_{\rm f}$ are the length of past and future motions. $T = T_{\rm p} + T_{\rm f}$ is the total length of motion. To facilitate subsequent explanation, we pad the past motion $\mathbb{X}^{-}$ to $\mathbb{X}\in \mathbb{R}^{T\times J \times 3}$ with zeros on future timestamps.
Our goal is to learn a prediction model $\mathcal{F}_{\rm pred}(\cdot)$ so that the predicted future motions $\widehat{\mathbb{X}}^+ = \mathcal{F}_{\rm pred}(\mathbb{X})$ are as close to the ground-truth future motions $\mathbb{X}^+$ as possible. 
To achieve an accurate prediction, one of the keys is capturing comprehensive spatial-temporal dependencies in the input motion. We are going to capture spatial-temporal dependencies via learning with auxiliary tasks.

\begin{figure*}[t]
\centering
\includegraphics[width=0.96\textwidth]{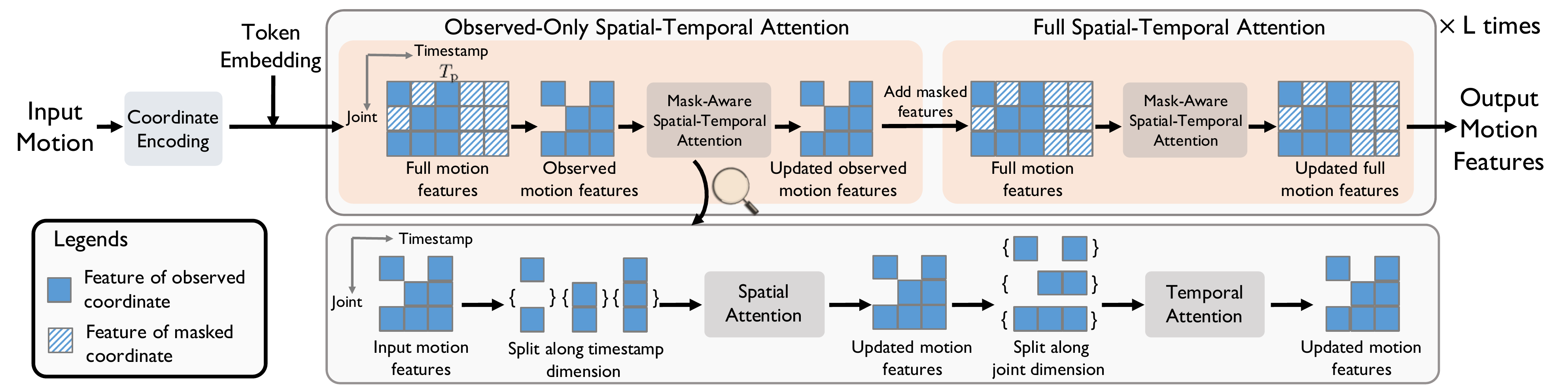}
\vspace{-2mm}
\caption{\small Auxiliary-adapted transformer (AuxFormer) as the dependency modeling network. 
We take the input of the masked prediction task as an example. 
The network first encodes coordinates and adds joint, timestamp, and masking information with token embedding. Then the network iteratively employs observed-only and full spatial-temporal attention to update the observed and whole motion features, respectively. Both observed-only and full spatial-temporal attention mechanisms use the same mask-aware spatial-temporal attention mechanism, which incorporates spatial and temporal attention for spatial and temporal dependency modeling.
}
\label{fig:network}
\vspace{-4mm}
\end{figure*}

\subsection{Framework Architecture}
Here we propose the model learning framework with multiple auxiliary tasks, which is sketched in Figure \ref{fig:framework}. 
The core idea of the framework is to learn the primary prediction task simultaneously with extra-designed auxiliary tasks that raise additional requirements to force the model to capture more comprehensive spatial-temporal dependencies. 
We introduce two additional auxiliary tasks: masking prediction and denoising. 
To be specific, taking the past human motion $\mathbb{X}\in \mathbb{R}^{T\times J\times 3}$ as the input, in the masking prediction task, each 3D coordinate $\mathbf{x}^t_j$ in the past motion has a probability of $p_{\rm m}$ to be masked by setting the value to zeros. The goal of the masking prediction task is to recover the masked coordinates from the remaining observed coordinates. 
In the denoising task, every 3D coordinate $\mathbf{x}^t_j$ has a probability of $p_{\rm n}$ to be noisy by adding a Gaussian noise $\epsilon \sim \mathcal{N}(0,\sigma)$, where $\sigma$ is the deviation. The goal of the denoising task is to erase the noise according to the remaining noiseless coordinates. 

To combine the two auxiliary tasks, we propose a joint learning framework that enables the primary future prediction task and the two auxiliary tasks to be learned simultaneously in an end-to-end manner. All three tasks share the same dependency modeling network but use different prediction heads.
Mathematically, the mask of masking prediction task $\mathbf{M}_{\rm M} \in \mathbb{R}^{T\times J}$ is defined as 
\begin{equation*}
    \setlength{\abovedisplayskip}{2pt}
   \setlength{\belowdisplayskip}{2pt}
    \mathbf{M}_{\rm M}(t,j) =
    \begin{cases}
        1 & \mathrm{if}\; t\leq T_{\rm p}\; \mathrm{and} \;\mathbf{x}_j^t\; \mathrm{is}\; \mathrm{unmasked}, \\
        0 & \mathrm{otherwise}.\\
    \end{cases}
\end{equation*}
Here we treat the future coordinates also be the "masked" coordinates since it is unknown. Similarly, we also define the mask of normal prediction task $\mathbf{M}_{\rm P}$ and the mask of denoising task $\mathbf{M}_{\rm D}$:
\begin{equation*}
    \setlength{\abovedisplayskip}{2pt}
   \setlength{\belowdisplayskip}{2pt}
    \mathbf{M}_{\rm P}(t,j) = \mathbf{M}_{\rm D}(t,j) =
    \begin{cases}
        1 & \mathrm{if}\; t\leq T_{\rm p},\; \\
        0 & \mathrm{otherwise}.\\
    \end{cases}
\end{equation*}
Let $\mathbb{X}$, $\mathbb{X}_{\rm M}$ and $\mathbb{X}_{\rm D}$ be the original input motion, input motion with masking and input motion with noise, 
the joint learning framework is formulated by,
\begin{equation}
    \setlength{\abovedisplayskip}{2pt}
   \setlength{\belowdisplayskip}{2pt}
\begin{aligned}
    &\mathbb{H}= \mathcal{F}(\mathbb{X},\mathbf{M}_{\rm P}), \widehat{\mathbb{X}} = \mathcal{P}_{\rm pred}(\mathbb{H}), \\ 
   &\mathbb{H}_{\rm M}=\mathcal{F}(\mathbb{X}_{\rm M},\mathbf{M}_{\rm M}), \widehat{\mathbb{X}}_{\rm M} = \mathcal{P}_{\rm mask}(\mathbb{H}_{\rm M}), \\
    &\mathbb{H}_{\rm D} = \mathcal{F}(\mathbb{X}_{\rm D},\mathbf{M}_{\rm D}),
    \widehat{\mathbb{X}}_{\rm D} = \mathcal{P}_{\rm denoise}(\mathbb{H}_{\rm D}), 
\end{aligned}
\label{eq:learning_framework}
\end{equation} 
where $\mathcal{F}(\cdot)$ denotes the auxiliary-adapted transformer (AuxFormer) as the dependency modeling network. $\mathbb{H}$, $\mathbb{H}_{\rm M}$, $\mathbb{H}_{\rm D} \in \mathbb{R}^{T\times J \times F}$ are motion features of future prediction task, mask prediction task and denoising task, respectively. $F$ is the dimension of the coordinate feature. $\mathcal{P}_{\rm pred}(\cdot)$, $\mathcal{P}_{\rm mask}(\cdot)$, and $\mathcal{P}_{\rm denoise}(\cdot)$ are three prediction heads, which are simple linear or MLP functions that map the $F$-dimensional space to the 3D coordinate space, and $\widehat{\mathbb{X}}$, $\widehat{\mathbb{X}}_{\rm M}$, $\widehat{\mathbb{X}}_{\rm D} \in \mathbb{R}^{T\times J \times 3}$ are the output sequences of the three tasks, which contains predicted future motions, predicted masked motions and denoised past motions, respectively. 

\vspace{-1mm}
Note that i) both the masking prediction task and the denoising task require the network to recover corrupted coordinates using their correlated coordinates according to their spatial-temporal dependencies. Through the two tasks, the dependency modeling framework is given extra force toward a more comprehensive spatial-temporal dependency modeling; 
ii) the dependency modeling network contains most of the model parameters and the prediction head is lightweight. Thus introducing extra auxiliary tasks into the learning framework only slightly increase the model size; 
iii) in the inference time we only perform the primary future prediction task by taking the future timestamps of the future prediction sequence as the predicted output $\widehat{\mathbb{X}}^{+} = \widehat{\mathbb{X}}_{[T_{\rm p}+1:T_{\rm f}]}$.

Compared to \cite{cui2021towards} that combines motion repairing with the motion prediction task, our framework has two major differences. First, \cite{cui2021towards} aims to predict motions from incomplete observations; while we consider the classical setting of motion prediction. Second, the motion repairing task in \cite{cui2021towards} is a necessary, non-splittable part of the model and will be included in the inference phase; while our auxiliary tasks serve as adjunctive, parallel branches to enhance model training and will not be involved in the inference phase.

\subsection{Loss Function}

For different tasks, we supervise different parts of the prediction sequences to satisfy the task demands. To be specific, for the prediction/masking/denoising task, we supervise the future/masked/past coordinates. Mathematically, assuming the masking set is $\mathcal{M}=\{(j,t)|\mathbf{M}(j,t)=0\}$, given the output sequence $\widehat{\mathbb{X}}$, $\widehat{\mathbb{X}}_{\rm M}$, $\widehat{\mathbb{X}}_{\rm D}$ of three tasks, the overall loss function is formulated by
\begin{equation*}
    \setlength{\abovedisplayskip}{-1pt}
   \setlength{\belowdisplayskip}{1pt}
\begin{aligned}
   & \mathcal{L}_{\rm prediction} = \frac{1}{T_{\rm f}J}\sum_{t=T_{\rm p}+1}^{T_{\rm f}}\sum_{j=1}^{J}\|\widehat{\mathbf{x}}_j^t - \mathbf{x}_j^t\|^2, \\ 
   \vspace{-1mm}
   & \mathcal{L}_{\rm mask} = \frac{1}{|\mathcal{M}|}\sum_{(j,t) \in \mathcal{M}} \|\widehat{\mathbf{x}}_{M,j}^t - \mathbf{x}_j^t\|^2, \\
   \vspace{-1mm}
   & \mathcal{L}_{\rm denoise} = \frac{1}{T_{\rm p}J}\sum_{t=1}^{T_{\rm p}}\sum_{j=1}^{J}\|\widehat{\mathbf{x}}_{D,j}^t - \mathbf{x}_j^t\|^2, 
   \\
   \vspace{-1mm}
    & \mathcal{L} = \mathcal{L}_{\rm prediction} + \alpha_1\mathcal{L}_{\rm mask} + \alpha_2\mathcal{L}_{\rm denoise}.
\end{aligned}
\end{equation*}
We use the average $\ell_2$ distance between different targets and predictions. $\alpha_1$ and $\alpha_2$ are weight hyperparameters. 

\section{Auxiliary-Adapted Transformer}
This section introduces an auxiliary-adapted transformer network to implement the dependency modeling network $\mathcal{F}(\cdot)$ in eq.\eqref{eq:learning_framework}. To work with auxiliary tasks, the network design faces two demands. First, it must learn spatial-temporal dependencies between corrupted and normal coordinates on a coordinate-wise basis to enable recovery. Second, the network must be adaptable to incomplete motions caused by the masking prediction task. To the first demand, we separately encode each 3D coordinate into an individual feature and use spatial-temporal attention to model coordinate-wise spatial-temporal dependencies. To the second demand, we add a masked token into the masked coordinate's feature to indicate it is masked and incorporate a mask-aware design into the spatial-temporal attention mechanism that enables inputs with arbitrary missing positions.

\subsection{Overview}
Mathematically, given the observed motion of an arbitrary task $\mathbb{X}\in \mathbb{R}^{T \times J \times 3}$ and its corresponding mask $\mathbf{M} \in \mathbb{R}^{T\times J}$ as the input, our auxiliary-adapted transformer works as,
\begin{subequations}
    \setlength{\abovedisplayskip}{2pt}
   \setlength{\belowdisplayskip}{2pt}
    \begin{align}
    &\mathbb{E} = \mathcal{F}_{\rm CE}(\mathbb{X}), \label{eq0}\\
      &  \mathbb{H} = \mathcal{F}_{\rm TE}(\mathbb{E},\mathbf{M}), \label{eq1}\\
&\mathbb{H}^{\prime} =\mathcal{F}^{(l)}_{\rm OSTA}(\mathbb{H},\mathbf{M}), \label{eq2}\\
&\mathbb{H}^{\prime\prime} =\mathcal{F}^{(l)}_{\rm FSTA}(\mathbb{H}^{\prime},\mathbf{1}), \label{eq3}
    \end{align}
\end{subequations}
where $\mathbb{E},\mathbb{H},\mathbb{H}^\prime,\mathbb{H}^{\prime\prime} \in \mathbb{R}^{T\times J \times F}$ are motion features, where $\mathbb{E}$ and $\mathbb{H}$ are initial and embedded features, while $\mathbb{H}^\prime$ and $\mathbb{H}^{\prime\prime}$ are updated features. $F$ is the feature dimension and $\mathbf{1}\in \mathbb{R}^{T\times J}$ is a mask matrix with all one values..

Step (\ref{eq0}) uses a coordinate encoder $\mathcal{F}_{\rm CE}(\cdot)$ to obtain the initial motion features. Step (\ref{eq1}) embeds joint, timestamp, and masked information into the initial coordinate features to obtain the embedded coordinate features using a token embedding function $\mathcal{F}_{\rm TE}(\cdot)$. Steps (\ref{eq2}) apply observed-only spatial-temporal attention $\mathcal{F}_{\rm OSTA}^{(l)}(\cdot)$ to update the observed coordinate features. Steps (\ref{eq3}) apply full spatial-temporal attention $\mathcal{F}_{\rm FSTA}^{(l)}(\cdot)$ to update the whole coordinate features. 
Step (\ref{eq2}) and (\ref{eq3}) are repeated iteratively $L$ times. 
The observed-only and full spatial-temporal attention have the same mask-aware spatial temporal attention mechanism with the only difference of the input mask.
The whole auxiliary-adapted transformer network is sketched in Figure \ref{fig:network}.  


Note that: i) instead of solely using full spatial-temporal attention, we additionally incorporate observed-only spatial-temporal attention. The intuition is modeling spatial-temporal dependencies inner observed coordinates provides more correlations to improve the ability to infer spatial-temporal dependencies between observed and masked coordinates; ii) instead of performing observed-only and full spatial-temporal attention separately for $L$ times, we employ an iterative approach because interaction with masked coordinate features can also enhance observed features. We evaluate the effects of these design choices in Section \ref{subsec:ablation}.

\subsection{Structure Details}
We now elaborate the details of each step.

\noindent\textbf{Coordinate encoding.}
Coordinate encoding aims to map each 3D coordinate to a high-dimensional embedding space for subsequent feature learning. Here we simply use a linear layer to implement the coordinate encoder $\mathcal{F}_{\rm CE}(\cdot)$.

\vspace{0.1cm}
\noindent\textbf{Token embedding.}
In token embedding, we encode the joint, timestamp and masking information. To encode
joint information, we consider dictionary learning to convert the joint number $j$ into a learnable code $\mathbf{w}_{{\rm J,j}} \in \mathbb{R}^{F}$, which represents the $j$th element of the joint embedding dictionary $\mathbf{W}_{{\rm J}} \in \mathbb{R}^{J \times F}$. Similarly, we convert the timestamp $t$ into a learnable code $\mathbf{w}_{{\rm T,t}} \in \mathbb{R}^{F}$. For the masked or future coordinates, we add a learnable masked token $\mathbf{w}_{\rm M} \in \mathbb{R}^{F}$. Mathematically, the embedded coordinate feature for the $j$th joint of the $t$th timestamp is
\begin{equation*}
    \setlength{\abovedisplayskip}{2pt}
   \setlength{\belowdisplayskip}{2pt}
\mathbf{h}_j^t =
\begin{cases}
\mathbf{e}_j^t + \mathbf{w}_{{\rm T},t}+ \mathbf{w}_{{\rm J},j} & {\rm if} \; \mathbf{M}(t,j) = 1,\\
\mathbf{w}_{\rm M} +  \mathbf{w}_{{\rm T},t}+ \mathbf{w}_{{\rm J},j} & {\rm if} \; \mathbf{M}(t,j)= 0.
\end{cases}
\end{equation*}
$\mathbf{e}^t_j$ is the initial feature of the $t$th timestamp of the $j$th joint.



\begin{table*}[t]
\setlength{\tabcolsep}{4.5pt}
\caption{\small Comparisons of short-term prediction on Human3.6M. Results at 80ms, 160ms, 320ms and 400ms in the future are shown. $\textbf{Bold}$/\underline{underline} font represent the best/second best result.}
\vspace{-3mm}
\renewcommand\arraystretch{0.85}
\resizebox{\textwidth}{!}{
\scriptsize
\begin{tabular}{c|cccc|cccc|cccc|cccc} \hline
Motion & \multicolumn{4}{c|}{Walking}                                   & \multicolumn{4}{c|}{Eating}                                    & \multicolumn{4}{c|}{Smoking}                                   & \multicolumn{4}{c}{Discussion}                                \\ \hline
millisecond        & 80ms          & 160ms         & 320ms         & 400ms         & 80ms          & 160ms         & 320ms         & 400ms         & 80ms          & 160ms          & 320ms          & 400ms          & 80ms           & 160ms          & 320ms          & 400ms          \\ \hline
Res-sup. \cite{martinez2017human}         & 29.4          & 50.8          & 76.0          & 81.5          & 16.8          & 30.6          & 56.9          & 68.7          & 23.0          & 42.6          & 70.1          & 82.7          & 32.9          & 61.2          & 90.9          & 96.2          \\
Traj-GCN \cite{mao2019learning}               & 12.3          & 23.0          & 39.8          & 46.1          & 8.4           & 16.9          & 33.2          & 40.7          &  7.9           &  16.2          & 31.9          & 38.9          & 12.5          & 27.4          & 58.5          & 71.7          \\
DMGNN \cite{li2020dynamic}            & 17.3          & 30.7          & 54.6          & 65.2          & 11.0          & 21.4          & 36.2          & 43.9          & 9.0           & 17.6          & 32.1          & 40.3          & 17.3          & 34.8          & 61.0          & 69.8          \\
MSRGCN \cite{dang2021msr}            &  12.2          &  22.7          &  38.6          &  45.2          &  8.4           & 17.1          &  33.0          &  40.4          & 8.0           & 16.3          &  31.3          &  38.2          &  12.0          &  26.8          &  57.1          &  69.7          \\
PGBIG \cite{ma2022progressively}             &   10.2          &   19.8          &    \underline{34.5}          &  \underline{40.3}          & \underline{7.0}           & 15.1         & 30.6          &   {38.1}          &   \underline{6.6}           &   {14.1}          &   {28.2}          &   {34.7}          &   \underline{10.0}          &   \underline{23.8}          &   \underline{53.6}          &   \underline{66.7}          \\ 
SPGSN \cite{li2022skeleton}             &   \underline{10.1}          &   \underline{19.4}          &   {34.8}          &   {41.5}          &   {7.1}           &   \underline{14.9}          &   \underline{30.5}          &   \underline{37.9}          &   {6.7}           &   \underline{13.8}          &   \underline{28.0}          &   \underline{34.6}          &   {10.4}          &   \underline{23.8}          &   \underline{53.6}          &   {67.1}          \\
AuxFormer (Ours) &  \textbf{{8.9}}&  \textbf{{16.9}}& \textbf{30.1}& \textbf{36.1}& \textbf{{6.4}}& \textbf{{14.0}}& \textbf{{28.8}}& \textbf{{35.9}}& \textbf{{5.7}}& \textbf{{11.4}}& {\textbf{22.1}}& \textbf{{27.9}}& \textbf{{8.6}}& \textbf{{18.8}}& \textbf{{38.8}}& \textbf{{49.2}}\\\hline
Motion           & \multicolumn{4}{c|}{Directions}                                & \multicolumn{4}{c|}{Greeting}                                  & \multicolumn{4}{c|}{Phoning}                                   & \multicolumn{4}{c}{Posing}                                    \\ \hline
millisecond        & 80ms          & 160ms         & 320ms         & 400ms         & 80ms          & 160ms         & 320ms         & 400ms         & 80ms          & 160ms          & 320ms          & 400ms          & 80ms           & 160ms          & 320ms          & 400ms          \\ \hline
Res-sup. \cite{martinez2017human}         & 35.4          & 57.3          & 76.3          & 87.7          & 34.5          & 63.4          & 124.6         & 142.5         & 38.0          & 69.3          & 115.0         & 126.7         & 36.1          & 69.1          & 130.5         & 157.1         \\
Traj-GCN \cite{mao2019learning}               & 9.0           & 19.9          & 43.4          &  53.7          & 18.7          & 38.7          & 77.7          & 93.4          & 10.2          & 21.0          & 42.5          & 52.3          & 13.7          & 29.9          &  66.6          &  84.1          \\
DMGNN \cite{li2020dynamic}             & 13.1          & 24.6          & 64.7          & 81.9          & 23.3          & 50.3          & 107.3         & 132.1         & 12.5          & 25.8          & 48.1          & 58.3          & 15.3          & 29.3          & 71.5          & 96.7          \\
MSRGCN \cite{dang2021msr}               &  8.6           &  19.7          &  43.3          & 53.8          &  16.5          &  37.0          &  77.3          &  93.4          &  10.1          &  20.7          &  41.5          &  51.3          &  12.8          &  29.4          & 67.0          & 85.0          \\
PGBIG \cite{ma2022progressively}               &   \underline{7.2}  &   {17.6} &   {40.9} &   \underline{51.5} &   {15.2} &   {34.1} &   {71.6} &   {87.1} &   \underline{8.3}  &   \underline{18.3} &   \underline{38.7} &   \underline{48.4} &   \underline{10.7} &   {25.7} &   {60.0} &   {76.6} \\ 
SPGSN \cite{li2022skeleton}              & 7.4          & \underline{17.2}          & \textbf{39.8}          & \textbf{50.3}          & \underline{14.6}         & \underline{32.6}          &  \underline{70.6}         &  \underline{86.4}         & 8.7 &\underline{18.3}& \underline{38.7}& 48.5        & \underline{10.7}& \underline{25.3}& \underline{59.9}& \underline{76.5}               \\ 
AuxFormer (Ours) &  \textbf{{6.8}}& \textbf{{17.0}}& \underline{40.3}& {51.6}& \textbf{{13.5}}& \textbf{{31.3}}&\textbf{69.2}&\textbf{85.4}& \textbf{{7.9}}& \textbf{{17.3}}& \textbf{{37.4}}& \textbf{{47.2}}& \textbf{{8.8}}& \textbf{{19.1}}& \textbf{{39.2}}& \textbf{{51.0}} \\ \hline
Motion           & \multicolumn{4}{c|}{Purchases}                                 & \multicolumn{4}{c|}{Sitting}                                   & \multicolumn{4}{c|}{Sittingdown}                               & \multicolumn{4}{c}{Takingphoto}                               \\ \hline
millisecond        & 80ms          & 160ms         & 320ms         & 400ms         & 80ms          & 160ms         & 320ms         & 400ms         & 80ms          & 160ms          & 320ms          & 400ms          & 80ms           & 160ms          & 320ms          & 400ms          \\ \hline
Res-sup. \cite{martinez2017human}         & 36.3          & 60.3          & 86.5          & 95.9          & 42.6          & 81.4          & 134.7         & 151.8         & 47.3          & 86.0          & 145.8         & 168.9         & 26.1          & 47.6          & 81.4          & 94.7          \\
Traj-GCN \cite{mao2019learning}              & 15.6          & 32.8          & 65.7          &  79.3          & 10.6          &  21.9          & 46.3          & 57.9          & 16.1          &  31.1          &  61.5          &  75.5          & 9.9           &  20.9          & 45.0          & 56.6          \\
DMGNN \cite{li2020dynamic}             & 21.4          & 38.7          & 75.7          & 92.7          & 11.9          & 25.1          & 44.6          &   \textbf{{50.2}}          & 15.0          & 32.9          & 77.1          & 93.0          & 13.6          & 29.0          & 46.0          & 58.8          \\
MSRGCN \cite{dang2021msr}               &  14.8          &  32.4          & 66.1          & 79.6          &  10.5          & 22.0          &  46.3          & 57.8          &  16.1          & 31.6          & 62.5          & 76.8          &  9.9           & 21.0          &  44.6          &  56.3         \\
PGBIG \cite{ma2022progressively}               &   \underline{12.5} &   {28.7} &    \textbf{{60.1}} &    \textbf{{73.3}} &   \underline{8.8}  &   \underline{19.2} &   {42.4} &  53.8 &   \underline{13.9} &   {27.9} &   \underline{57.4} &   \underline{71.5} &   \underline{8.4}  &   \underline{18.9} &   \underline{42.0} &   {53.3} \\ 
SPGSN \cite{li2022skeleton}               & 12.8 &\underline{28.6} &\underline{61.0}& \underline{74.4}         & 9.3& 19.4 & \underline{42.3}&  {53.6}          & 14.2 &\underline{27.7}& \textbf{56.8}& \textbf{70.7}       & 8.8& \underline{18.9}&  \textbf{{41.5}}&  \textbf{{52.7}}              \\ 
AuxFormer (Ours) & \textbf{{11.9}} & \textbf{{28.0}} &{61.8} &76.3& \textbf{{8.7}}&  \textbf{{19.0}} & \textbf{42.1}& \underline{53.3}& \textbf{{13.5}}&  \textbf{{27.6}}&  {57.7}&  {72.2}& \textbf{{8.2}}&  \textbf{{18.4}}& \textbf{41.5}& \underline{53.0} \\\hline
Motion           & \multicolumn{4}{c|}{Waiting}                                   & \multicolumn{4}{c|}{Walking Dog}                                & \multicolumn{4}{c|}{Walking Together}                           & \multicolumn{4}{c}{Average}                                       \\ \hline
millisecond       & 80ms          & 160ms         & 320ms         & 400ms         & 80ms          & 160ms         & 320ms         & 400ms         & 80ms          & 160ms          & 320ms          & 400ms          & 80ms           & 160ms          & 320ms          & 400ms          \\ \hline
Res-sup. \cite{martinez2017human}         & 30.6          & 57.8          & 106.2         & 121.5         & 64.2          & 102.1         & 141.1         & 164.4         & 26.8          & 50.1          & 80.2          & 92.2          & 34.7          & 62.0          & 101.1         & 115.5         \\
Traj-GCN \cite{mao2019learning}               & 11.4          & 24.0          & 50.1          & 61.5          & 23.4          & 46.2          & 83.5          & 96.0          &  10.5          & 21.0          & 38.5          & 45.2          & 12.7          & 26.1          & 52.3          & 63.5          \\
DMGNN \cite{li2020dynamic}             & 12.2          & 24.2          & 59.6          & 77.5          & 47.1          & 93.3          & 160.1         & 171.2         & 14.3          & 26.7          & 50.1          & 63.2          & 17.0          & 33.6          & 65.9          & 79.7          \\
MSRGCN \cite{dang2021msr}               &  10.7          &  23.1          &  48.3          & 59.2          &  20.7          &  42.9          &  80.4          &  93.3          & 10.6          &  20.9          &  37.4         &  43.9          &  12.1          &  25.6          &  51.6          &  62.9         \\

PGBIG \cite{ma2022progressively}               &   \underline{8.9}  &   {20.1} &   {43.6} &   {54.3} &   {18.8} &   {39.3} &   {73.7} &   {86.4} &   \underline{8.7}  &   {18.6} &   {34.4} &   {41.0} &   \underline{10.3}  &   {22.7}  &   {47.4}  &   {58.5} \\ 
SPGSN \cite{li2022skeleton}               & 9.2& \underline{19.8}& \underline{43.1}& \underline{54.1}        &  \underline{17.8}& \underline{37.2}& \underline{71.7}& \underline{84.9}               & 8.9 &\underline{18.2}& \underline{33.8}& \underline{40.9}        & 10.4 &\underline{22.3} &\underline{47.1}& \underline{58.3}         \\ 
AuxFormer (Ours) & \textbf{{8.2}}&  \textbf{{18.5}}&  \textbf{{41.2}}&  \textbf{{52.2}}&  \textbf{{17.1}}& \textbf{{36.5}} & \textbf{70.4}&  \textbf{{83.0}}& \textbf{7.8} & \textbf{15.9}&  \textbf{30.2}&  \textbf{{37.0}}&  \textbf{9.5} & \textbf{20.6} & \textbf{43.4}&  \textbf{{54.1}}\\\hline
\end{tabular}
}
\label{tab:human3.6_shortterm}
\vspace{-3mm}
\end{table*}
\begin{table*}[t]
\caption{\small Comparisons of long-term prediction on 7 representative actions on Human3.6M. Results at 560ms and 1000ms in the future are shown. $\textbf{Bold}$/\underline{underline} font represent the best/second best result.}
\vspace{-0.3cm}
\renewcommand\arraystretch{0.9}
\resizebox{\textwidth}{!}{
\begin{tabular}{c|cc|cc|cc|cc|cc|cc|cc|cc} \hline
Motion & \multicolumn{2}{c|}{Walking}    & \multicolumn{2}{c|}{Smoking}     & \multicolumn{2}{c|}{Discussion}     & \multicolumn{2}{c|}{Greeting}  & \multicolumn{2}{c|}{Posing} & \multicolumn{2}{c|}{Walking Dog}    & \multicolumn{2}{c|}{Walking Together}         & \multicolumn{2}{c}{Average}      \\ \hline
millisecond        & 560ms         & 1000ms         & 560ms         & 1000ms         & 560ms          & 1000ms         & 560ms          & 1000ms         & 560ms         & 1000ms         & 560ms          & 1000ms         & 560ms            & 1000ms           & 560ms          & 1000ms         \\ \hline
Res-Sup. \cite{martinez2017human} & 81.7 & 100.7 & 94.8 & 137.4 & 121.3 & 161.7 & 156.3 & 184.3 & 165.7 & 236.8 & 173.6 & 202.3 & 94.5 & 110.5 & 129.2 & 165.0\\
Traj-GCN \cite{mao2019learning} &54.1 & 59.8 & 50.7 & 72.6 & 91.6 & 121.5 & 115.4 & 148.8 & 114.5 & 173.0 & 111.9 & 148.9 & 55.0 & 65.6 & 81.6 & 114.3 \\
DMGNN \cite{li2020dynamic} & 71.4 & 85.8 & 50.9 & 72.2 & 81.9 & 138.3 & 144.5 & 170.5 & 125.5 & 188.2 & 183.2 & 210.2 & 70.5 & 86.9 & 93.6 & 127.6 \\
MSRGCN \cite{dang2021msr} &   {52.7}          &   {63.0}           & 49.5 & 71.6 & 88.6 & \underline{117.6} & 116.3 & 147.2 & 116.3 & 174.3 & 111.9 & 148.2 & 52.9 & 65.9 & 81.1 & 114.2 \\
PGBIG \cite{ma2022progressively} &   {48.1} & {56.4} & \underline{46.5} & 69.5 & \underline{87.1} & 118.2 & \textbf{110.2} & 143.5 & \underline{106.1} & \underline{164.8}  & 104.7 & 139.8 & 51.9 & 64.3 & \underline{76.9} & 110.3 \\ 
SPGSN \cite{li2022skeleton} &\underline{46.9} &\underline{53.6}& 46.7 & \underline{68.6} & 89.7 & 118.6 & 111.0 & \underline{143.2} & 110.3 & 165.4 & \textbf{102.4} & \underline{138.0} & \underline{49.8} & \underline{60.9} & 77.4 & \underline{109.6}\\
AuxFormer (Ours)  & \textbf{43.8} & \textbf{52.0} & \textbf{42.0} & \textbf{63.0} & \textbf{77.6} & \textbf{102.3} & \underline{110.5} & \textbf{141.6} & \textbf{91.6} & \textbf{137.1} & \underline{103.3} & \textbf{133.3} & \textbf{47.3} & \textbf{58.8} & \textbf{75.3} & \textbf{107.0} \\\hline
\end{tabular} 
}
\label{tab:Human3.6long-term}
\vspace{-2mm}
\end{table*}


\noindent\textbf{Mask-aware spatial-temporal attention.}
The mask-aware spatial-temporal attention is used to model spatial-temporal dependencies among assigned features based on the input mask, which implements $\mathcal{F}_{\rm OSTA}^{(l)}(\cdot)$ and $\mathcal{F}_{\rm FSTA}^{(l)}(\cdot)$ in eq.(\ref{eq2}) and eq.(\ref{eq3}) respectively. As the spatial relationship between joints and the temporal relation between timestamps have different patterns, we use spatial and temporal attention mechanisms to model the spatial and temporal dependencies separately. 
Spatial attention considers features of the same timestamp, while temporal attention considers features of the same joint. 


The spatial attention simultaneously operates features at different timestamps. For the feature $\mathbf{H}^t \in \mathbb{R}^{J \times D}$ of $t$th timestamp, we first compute its spatial attention matrix $\mathbf{A}_s \in \mathbb{R}^{J\times J}$ according to the input mask,
\begin{equation*}
    \setlength{\abovedisplayskip}{2pt}
   \setlength{\belowdisplayskip}{2pt}
    \mathbf{A}_s(j_1,j_2) =
    \begin{cases}
    1 & \mathrm{if} \; \mathbf{M}(t,j_1)=1 \; \mathrm{and}\; \mathbf{M}(t,j_2)=1 \\
    0 & \mathrm{otherwise}, \\
    \end{cases}
\end{equation*}
where $\mathbf{A}_s(j_1,j_2)$ indicates whether performing attention between joint $j_1$ and $j_2$. Then the spatial attention is 
\begin{equation*}
    \setlength{\abovedisplayskip}{2pt}
   \setlength{\belowdisplayskip}{2pt}
\begin{aligned}
       & \mathbf{Q}^t,\mathbf{K}^t,\mathbf{V}^t = f_{\rm QKV}(\mathbf{H}^t), \\
    &\mathrm{\mathbf{O}^t} = \big[\mathrm{Softmax}(\frac{\mathbf{Q}^t {\mathbf{K}^t}^\top}{\sqrt{F}})\cdot \mathbf{A}_s\big]\mathbf{V}^t,\\
    &\mathbf{H}^t \leftarrow  \mathbf{H}^t + \mathrm{FFN}(\mathop{||}_{i=1}^H \mathrm{\mathbf{O}}_i^t ),
\end{aligned}
\end{equation*}
where $f_{\rm QKV}(\cdot)$ are linear operations, $\mathrm{Softmax}(\cdot)$ represents the softmax function. We use a regular multi-head feed-forward operation that produces totally $H$ heads $\{\mathbf{O}_i^t\}$ in parallel and use a feedforward function $\mathrm{FFN}(\cdot)$ to obtain the output of spatial attention. $\mathop{||}$ denotes concatenation. 

Similarly, temporal attention simultaneously operates features of different joints. For the $j$th joint's feature $\mathbf{H}_j \in \mathbb{R}^{T \times D}$, we compute its temporal attention matrix $\mathbf{A}_t \in \mathbb{R}^{T\times T}$ by
\begin{equation*}
    \setlength{\abovedisplayskip}{2pt}
   \setlength{\belowdisplayskip}{2pt}
    \mathbf{A}_t(t_1,t_2) =
    \begin{cases}
    1 & \mathrm{if} \; \mathbf{M}(j,t1)=1, \; \mathrm{and}\; \mathbf{M}(j,t_2)=1, \\
    0 & \mathrm{otherwise}. \\
    \end{cases}
\end{equation*}
The temporal attention is 
\begin{equation*}
    \setlength{\abovedisplayskip}{2pt}
   \setlength{\belowdisplayskip}{2pt}
\begin{aligned}
       & \mathbf{Q}_j,\mathbf{K}_j,\mathbf{V}_j = f_{\rm QKV}(\mathbf{H}_j) , \\
    &\mathrm{\mathbf{O}_j} = \big[\mathrm{Softmax}(\frac{\mathbf{Q}_j {\mathbf{K}_j}^\top}{\sqrt{D}})\cdot \mathbf{A}_t\big]\mathbf{V}_j,\\
    &\mathbf{H}_j \leftarrow  \mathbf{H}_j + \mathrm{FFN}(\mathop{||}_{i=1}^H \mathrm{\mathbf{O}}_{j,i} ).
\end{aligned}
\end{equation*}


Compared to the previous spatial-temporal transformer structure used in \cite{aksan2021spatio}, our auxiliary-adapted transformer has two advantages: adaptability to missing data cases, which the previous method cannot handle, and modeling of global spatial-temporal dependencies between arbitrary timestamps, while previous method only models temporal dependencies within a time window.

\begin{table*}[t]
\setlength{\tabcolsep}{6pt}
    \centering
    \caption{\small Prediction MPJPEs of methods on CMU Mocap for both short-term and long-term prediction across 7 actions, as well as their average prediction results all on actions. $\textbf{Bold}$/\underline{underline} font represent the best/second best result.}
    \vspace{-2mm}
    \footnotesize
    \renewcommand{\arraystretch}{0.95}
    \resizebox{1\textwidth}{!}{
        \begin{tabular}{c|ccccc|ccccc|ccccc|ccccc}
        \hline
        Motion & \multicolumn{5}{c|}{Basketball} & \multicolumn{5}{c|}{Basketball Signal} & \multicolumn{5}{c|}{Jumping} & \multicolumn{5}{c}{Running} \\ 
        millisecond & 80 & 160 & 320 & 400 & 1000 & 80 & 160 & 320 & 400 & 1000 & 80 & 160 & 320 & 400 & 1000 & 80 & 160 & 320 & 400 & 1000 \\ \hline
        Res-sup.~\cite{martinez2017human} & 15.45 & 26.88 & 43.51 & 49.23 & 88.73 & 20.17 & 32.98 & 42.75 & 44.65 & 60.57 &  26.85 & 48.07 & 93.50 & 108.90 & 162.84 & 25.76 &48.91 &88.19& 100.80 &158.19\\
        DMGNN~\cite{li2020dynamic} & 15.57 & 28.72 & 59.01 & 73.05 & 138.62 & 5.03 & 9.28 & 20.21 & 26.23 & 52.04  & 31.97 & 54.32 & 96.66 & 119.92 & 224.63 &17.42 &26.82 &38.27 &40.08 &\underline{46.40} \\
        Traj-GCN~\cite{mao2019learning} & 11.68 & 21.26 & 40.99 & 50.78 & 97.99 & 3.33 & 6.25 & 13.58 & 17.98 & 54.00  & 17.18 & 32.37 & 60.12 & 72.55 & 127.41 & 14.53 &24.20 &37.44 &41.10 &51.73 \\
        MSR-GCN~\cite{dang2021msr} & 10.28 & 18.94 & { 37.68} & { 47.03} & { 86.96} & 3.03 & 5.68 & 12.35 & 16.26 & \underline{47.91}  & 14.99 & 28.66 & { 55.86} & { 69.05} & \underline{ 124.79} &12.84& 20.42& 30.58 &34.42 &48.03\\
        STSGCN~\cite{sofianos2021space} & 12.56 & 23.04 & 41.92 & 50.33 & 94.17 & 4.72 & 6.69 & 14.53 & 17.88 & 49.52  & 17.52 & 31.48 & 58.74 & 72.06 & 127.40 & 16.70 &27.58& 36.15& 36.42& 55.34 \\
        PGBIG~\cite{ma2022progressively} & \underline{9.53} &\underline{17.53}& \textbf{35.32} &\textbf{44.23}& \underline{84.14}&\underline{2.71}& \underline{4.88}& \underline{10.77}& \underline{14.63} & 50.19&\underline{13.93}& \underline{27.78}& \underline{55.80} &\underline{69.01} &125.60 &12.69 & 23.18 & 38.31&42.24&51.71\\
        SPGSN \cite{li2022skeleton}& {10.24} & {18.54} & 38.22 & 48.68 & 89.58 & {2.91} & {5.25} & {11.31} & {15.01} & \textbf{{47.31}}  & {14.93} & {28.16} & 56.72 & 71.16 & 125.20 &\underline{10.75} &\underline{16.67}& \underline{26.07}& \underline{30.08}& 52.92\\
        AuxFormer (Ours) & \textbf{9.35} & \textbf{17.06}&\underline{35.46} &\underline{45.50}& \textbf{80.77}&\textbf{2.62}&\textbf{4.79}&\textbf{10.57}&\textbf{14.20}& 48.19&\textbf{12.79}&\textbf{25.52}&\textbf{53.37}&\textbf{67.28} & \textbf{124.34} &
           \textbf{9.98} & \textbf{15.78} & \textbf{25.31} &\textbf{28.81}  & \textbf{41.64} \\  \hline
        Motion & \multicolumn{5}{c|}{Soccer} & \multicolumn{5}{c|}{Walking} & \multicolumn{5}{c|}{Washing Window} & \multicolumn{5}{c}{Average} \\ 
        millisecond & 80 & 160 & 320 & 400 & 1000 & 80 & 160 & 320 & 400 & 1000 & 80 & 160 & 320 & 400 & 1000 & 80 & 160 & 320 & 400 & 1000 \\ 
        \cline{1-21}
        Res-sup.~\cite{martinez2017human} & 17.75 & 31.30 & 52.55 & 61.40 & 107.37 & 44.35 & 76.66 & 126.83 & 151.43 & 194.33 & 22.84 & 44.71 & 86.78 & 104.68 & 202.73 & 24.74 & 44.21 &  76.30 & 88.73 & 139.30\\
        DMGNN~\cite{li2020dynamic} & 14.86 & 25.29 & 52.21 & 65.42 & 111.90 & 9.57 & 15.53 & 26.03 & 30.37 & 67.01 & 7.93 & 14.68 & 33.34 & 44.24 & 82.84 & 14.07 & 24.44 & 45.90 & 55.45 & 104.33 \\
        Traj-GCN~\cite{mao2019learning} & 13.33 & 24.00 & 43.77 & 53.20 & 108.26 & 6.62 & 10.74 & 17.40 & 20.35 &  \underline{34.41} & 5.96 & 11.62 & 24.77 & 31.63 & 66.85 & 9.94 & 18.02 & 33.55 & 40.95 & 81.85\\
        MSR-GCN~\cite{dang2021msr} & 10.92 & 19.50 & 37.05 & 46.38 & \underline{ 99.32} & { 6.31} & 10.30 & 17.64 & 21.12 & 39.70 & 5.49 & 11.07 & 25.05 & 32.51 & 71.30 & 8.72 & 15.83 & 30.57 & 38.10 & 79.01\\
        STSGCN~\cite{sofianos2021space} & 13.49 & 25.24 & 39.87 & 51.58 & 109.63 & 7.18 & 10.99 & 17.84 & 22.61 & 44.12 & 6.79 & 12.10 & 24.92 & 36.66 & 69.48 & 10.80 & 18.19 & 31.18 & 41.05 & 81.76\\
        PGBIG \cite{ma2022progressively} & 11.09& 20.62& 39.48& 48.72& 99.98&\underline{6.23} &10.34 &16.84 &\underline{19.76}& \textbf{33.92}&\textbf{4.63}& \textbf{9.16}& \textbf{20.87}& \textbf{27.34}&\underline{65.69} &\underline{8.20} &15.41& 30.13 &37.27&\underline{76.69}\\
        SPGSN \cite{li2022skeleton}& \underline{10.86} & \underline{18.99} & \textbf{{35.05}} & \textbf{{45.16}} & 99.51 & 6.32 & \underline{10.21} & \underline{16.34} & {20.19} & 34.83 & {4.86} & {9.44} & \underline{21.50} & \underline{28.37} & \textbf{{65.08}} & {8.30} & \underline{14.80} & \underline{28.64} & \underline{36.96} & {77.82}\\ 
            AuxFormer (Ours) & \textbf{10.01}& \textbf{18.21}&\underline{36.31}&\underline{45.79}&\textbf{95.98}&\textbf{5.76}&\textbf{9.16}&\textbf{15.69}&\textbf{18.80}&34.81&\underline{4.69}&\underline{9.39}&21.87&28.83& 72.90& \textbf{7.54}&\textbf{13.78}&\textbf{27.95}&\textbf{35.39} &\textbf{76.32} \\
                \hline
    \end{tabular}}
    \vspace{-5mm}
    \label{tab:pred_cmu}
\end{table*}

\section{Experiment}

\subsection{Datasets}
\vspace{-1mm}
\noindent \textbf{Human3.6M} \cite{ionescu2013human3} has 7 subjects performing 15 types of actions with 22 body joints. The sequences are downsampled temporally by 2 and converted to 3D coordinates, excluding global rotation and translation of the pose. Subjects S5 and S11 are reserved for testing and validation, respectively, while the remaining subjects are used for training.

\vspace{0.1cm}
\noindent \textbf{CMU-MoCap} has 8 human action types including 38 body joints which are also converted to 3D coordinates. The global rotations and translations of the poses are excluded. Following \cite{Dang_2021_ICCV,mao2019learning}, we keep 25 joints and divide the training and testing datasets. 

\vspace{0.1cm}
\noindent\textbf{3DPW} \cite{von2018recovering} is a dataset for human pose prediction containing indoor and outdoor activities. They are represented in the 3D space and each subject has 23 joints. The frame rate of the motions is 30Hz.

\begin{table}[t]
      \small 
      \centering
      \caption{\small The average prediction MPJPEs on 3DPW dataset. $\textbf{Bold}$/\underline{underline} font represent the best/second best result.}
      \vspace{-3mm}
      \renewcommand{\arraystretch}{1.0}
      \resizebox{1.0\columnwidth}{!}{
      \begin{tabular}{|c|cccccc|}
          \hline
          ~ & \multicolumn{6}{c|}{Average MPJPE} \\
          \hline
          millisecond & 100 & 200 & 400  & 600 & 800  & 1000 \\
          \hline
          Res-sup.~\cite{martinez2017human} & 102.28 & 113.24 & 173.94  & 191.47& 201.39  & 210.58 \\
          CSM~\cite{li2018convolutional} & 57.83 & 71.53 & 124.01  & 155.16 & 174.87  & 187.06 \\
          Traj-GCN~\cite{mao2019learning} & 16.28 & 35.62 & 67.46 &90.36& 106.79  & 117.84 \\
          DMGNN~\cite{li2020dynamic} & 17.80 & 37.11 & 70.38 & 94.12 & 109.67  & 123.93  \\
          HisRep~\cite{mao2020history} & 15.88 & 35.14 & 66.82  & 93.55 & 107.63 & 114.75 \\
          MSR-GCN~\cite{dang2021msr} & 15.70 & 33.48 & 65.02 & 93.81& 108.15 & 116.31 \\
          STSGCN~\cite{sofianos2021space} & 18.32 & 37.79 & 67.51& 92.75& 106.65  & 112.22 \\
          PGBIG \cite{ma2022progressively} & 17.66 & 35.32 & 67.83 & \underline{89.60} & \underline{102.59} & \underline{109.41} \\
          SPGSN \cite{li2022skeleton} & \underline{15.39} & \underline{32.91} & \underline{64.54} &{91.62} & {103.98}  & {111.05} \\
                    \hline
        AuxFormer (Ours) &\textbf{14.21}& \textbf{30.04} &\textbf{58.50} & \textbf{89.45}& \textbf{100.78} & \textbf{107.45} \\
          \hline
      \end{tabular}}
      \vspace{-4mm}
      \label{tab:pred_3DPW}
\end{table}

\subsection{Model and Experimental Settings}
\noindent\textbf{Implementation details.} We set the feature dimension $F$ as 64/96/80 in Human3.6M/CMU Mocap/3DPW dataset. For the short/long-term motion prediction, the number of attention layers $L$ is 3/4 in the dependency modeling network.
We use 8 heads in the multi-head attention. We use a mask ratio of 50$\%$ and a noisy ratio of 30$\%$. 
The loss weight $\alpha_1,\alpha_2$ are both set to 1. The model is trained for 50 epochs with a batch size of 16. We use the Adam optimizer to train the model on a single NVIDIA RTX-3090 GPU. We scale the input motion by 1/100 and rescale the output motion.
To obtain a more generalized evaluation with lower test bias, we use all the clips in the 5th subject of H3.6M and the test folder of CMU Mocap, instead of testing on a few samples like in \cite{gui2018adversarial,li2018convolutional,li2020dynamic}. For more implementation details, please refer to the supplementary material.


\vspace{0.1cm}
\noindent\textbf{Evaluation Metrics.} We use the Mean Per Joint Position Error (MPJPE), which calculates the average $\ell$2 distance between predicted joints and target ones at each prediction timestamp in 3D Euclidean space. 
Compared to previous mean angle error (MAE) \cite{li2020dynamic,martinez2017human}, the MPJPE reflects larger degrees of freedom of human poses and covers larger ranges of errors for a clearer comparison.

\subsection{Main Results}
\noindent\textbf{Short-term prediction.} 
Short-term motion prediction aims to predict the poses within 400 milliseconds. 
Table \ref{tab:human3.6_shortterm} presents the MPJPEs of our method and many previous methods on all actions at multiple prediction timestamps on the H3.6M dataset. We see that i) our method achieves superior performance at most of the timestamps and very close performance to the best results on other timestamps; ii) our method achieves significantly lower MPJPEs on average by $8.7\%/7.6\%/7.9\%/7.2\%$ at 80/160/320/400ms. 

Besides, Table \ref{tab:pred_cmu} reports the comparison of our method and many previous methods on the CMU Mocap dataset. We show our method achieves more effective prediction and has a significant improvement by $9.2\%/6.9\%/3.3\%/3.7\%$ at 80/160/320/400ms. In Table \ref{tab:pred_3DPW}, we present the average MPJPEs across all the test samples at different prediction timestamps on the 3DPW dataset  Compared to the state-of-the-art methods, our method greatly reduces the MPJPE by $7.7\%/8.7\%/9.4\%$ at 100/200/400ms.

\begin{figure}[t]
\centering
\includegraphics[width=0.48\textwidth]{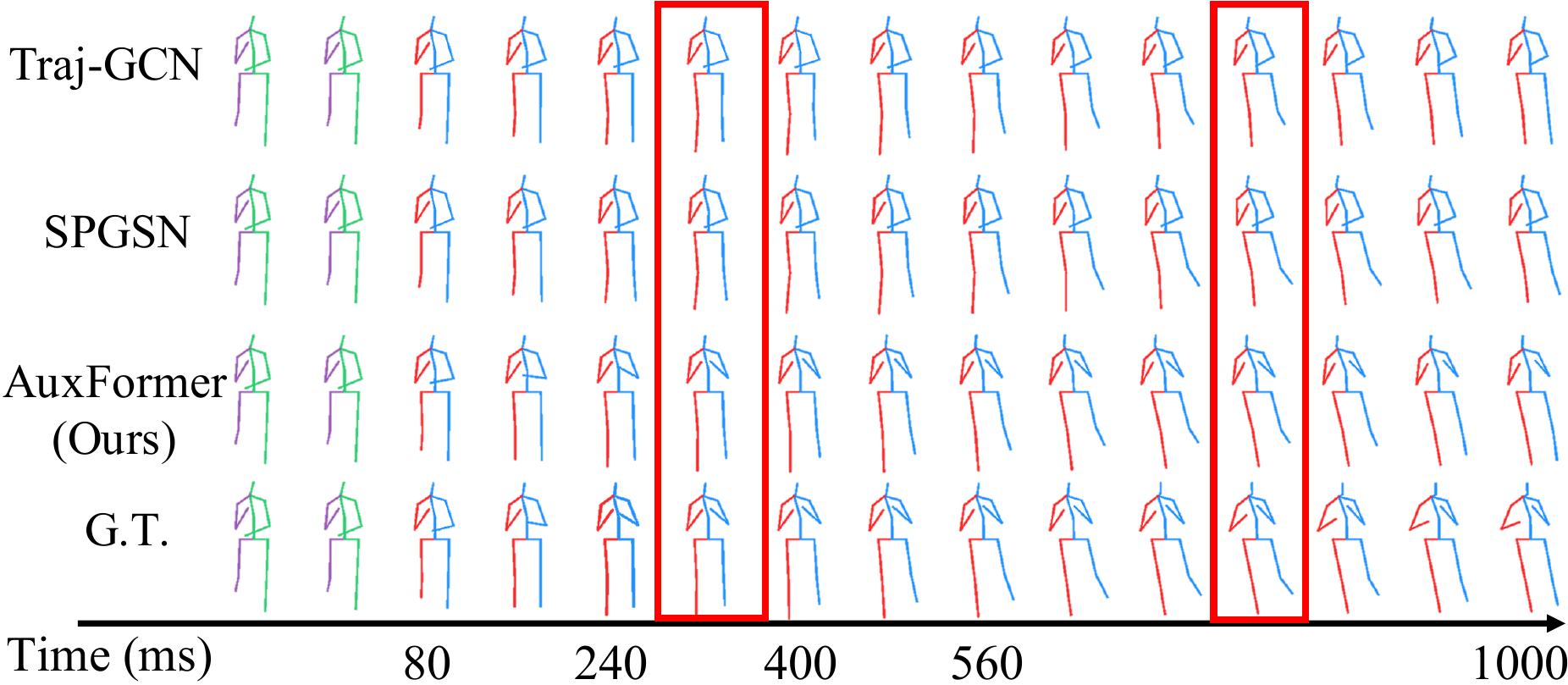}
\vspace{-6mm}
\caption{\small Visualization results of different methods on H3.6M.}
\label{fig:vis}
\vspace{-4mm}
\end{figure}



\vspace{0.1cm}
\noindent\textbf{Long-term prediction.}
Long-term motion prediction aims to predict the poses over 400 milliseconds. Table \ref{tab:Human3.6long-term} shows prediction MPJPEs of various methods at 560 and 1000 ms. Due to the space limit, we show the results of seven representative actions and the average results of all actions. More detailed tables are provided in the supplementary material. We see that our method achieves a more accurate prediction on most of the actions and our method has a lower MPJPE by $2.1\%/2.4\%$ at 560/1000ms compared to state-of-the-art methods.
Table \ref{tab:pred_cmu} and \ref{tab:pred_3DPW} also report the comparison of our method with previous methods on the CMU Mocap and 3DPW datasets. We see that our method also significantly outperforms the state-of-the-art method on the CMU Mocap and the 3DPW dataset.


\vspace{0.1cm}
\noindent\textbf{Qualitative results.}
Figure \ref{fig:vis} shows an example of the predicted poses of different methods. We can see our prediction result is more accurate than two representative baselines, especially on the motion of two arms.

\subsection{Ablation Study and Model Analysis}
\label{subsec:ablation}
\noindent\textbf{Effect of auxiliary tasks.}
We investigate the impact of two auxiliary tasks in our learning framework: masking prediction (Mask) and denoising (Denoise). Table \ref{table:abltion_operation1} and \ref{table:abltion_operation2} show the results on the H3.6M and CMU Mocap datasets, respectively. "Pred" denotes the future prediction task. We observe that i) adding the masking prediction and denoising tasks individually leads to a significant improvement in the model performance, demonstrating the effectiveness of these auxiliary tasks; ii) combining the two auxiliary tasks further imporves the model's performance.

\begin{table}[t]
  \centering
\renewcommand\arraystretch{0.9}
    \setlength{\tabcolsep}{2.5pt}
  \footnotesize
  \caption{\small Ablation study on auxiliary tasks on H3.6M.}
    \vspace{-3mm}
  \footnotesize
    \begin{tabular}{ccc|ccccccc}
    \toprule
        Pred & Mask & Denoise  & 80ms & 160ms & 320ms & 400ms & 560ms & 1000ms  \\
    \midrule
   \checkmark & & & 10.2 & 21.8 & 45.7 & 57.0 & 79.7& 110.5 \\
      \checkmark&\checkmark & &  9.7 &21.0 & 44.0& 54.8 & 77.7&108.2\\
       \checkmark& &\checkmark & 9.6 & 20.9& 44.2& 55.0 & 76.6 & 108.1  \\
       \checkmark& \checkmark&\checkmark  & \textbf{9.5} &\textbf{20.6}& \textbf{43.4} &\textbf{54.1}&\textbf{75.3}&\textbf{107.0} \\
    \bottomrule
    \end{tabular}%
  \label{table:abltion_operation1}%
\vspace{-2mm}
\end{table}

\begin{table}[t]
  \centering
    \renewcommand\arraystretch{0.9}
    \setlength{\tabcolsep}{3pt}
  \footnotesize
  \caption{\small Ablation study on auxiliary tasks on CMU Mocap.}
    \vspace{-3mm}
  \small
    \begin{tabular}{ccc|cccccc}
    \toprule
        Pred & Mask & Denoise  & 80ms & 160ms & 320ms & 400ms & 1000ms \\
    \midrule
   \checkmark & & & 8.57 & 15.48 & 30.40 & 38.35& 75.94  \\
      \checkmark&\checkmark & &  8.25&14.78&29.13&36.46 & 74.13\\
       \checkmark& &\checkmark &  8.14&14.57 &28.98 & 36.60 & 72.95 \\
       \checkmark& \checkmark&\checkmark  & \textbf{7.89} &\textbf{14.27}& \textbf{28.36} &\textbf{35.60} & \textbf{71.23} \\
    \bottomrule
    \end{tabular}%
  \label{table:abltion_operation2}%
  \vspace{-2mm}
\end{table}

\begin{figure}[t]
\centering
\includegraphics[width=0.46\textwidth]{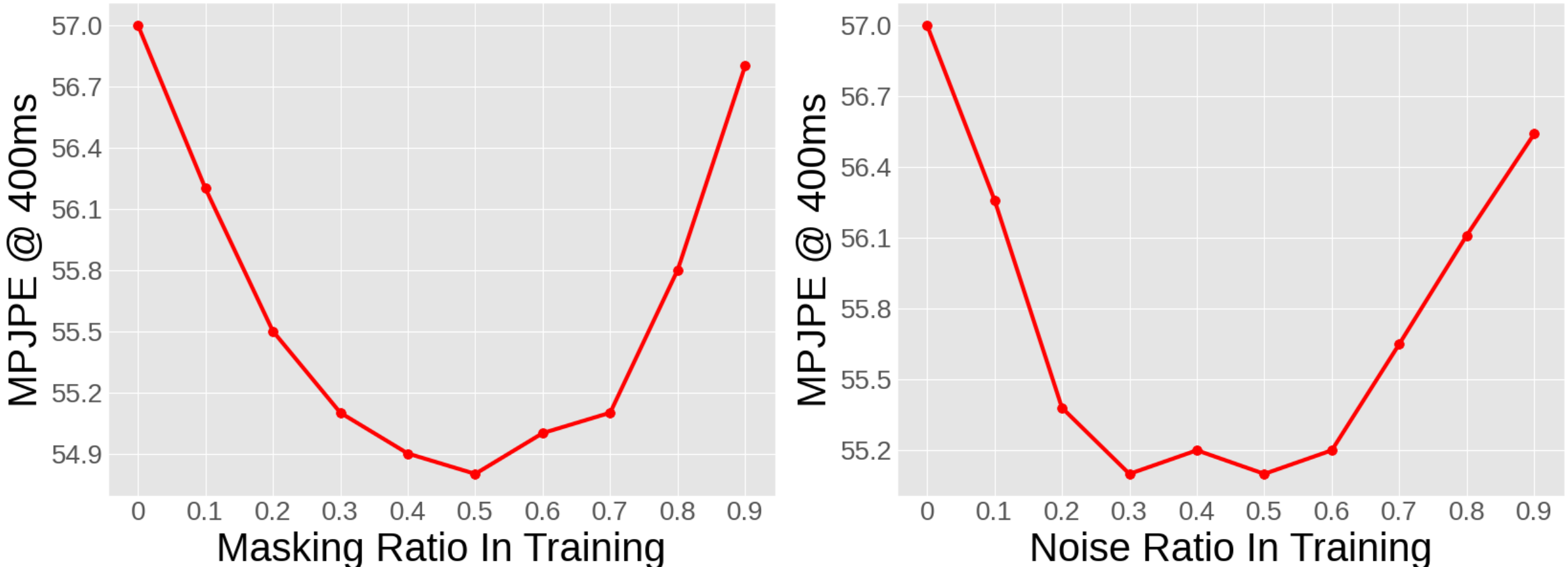}
\vspace{-2.5mm}
\caption{\small Effect of different masking ratios and noise ratios in the model training on H3.6M.}
\label{fig:ablation_mask}
\vspace{-2mm}
\end{figure}

\vspace{0.1cm}
\noindent\textbf{Effect of masking \& noise ratios.} We explore the effect of different masking ratios and noise ratios during model training. Figure \ref{fig:ablation_mask} shows the results. The noise deviation is set to 50. We see that i) setting masking and noising ratios that are either too low or too high results in suboptimal performance, as the auxiliary tasks become either too easy or too difficult, rendering them unsuitable for model learning; ii) a moderate masking and noising ratio yielded the best results, with an appropriate masking ratio range of 0.3 to 0.7 and an appropriate noising ratio range of 0.2 to 0.6.

\vspace{0.1cm}
\noindent\textbf{Ablation on model structure.}
We explore the effect of designs in our dependency modeling network structure. Table \ref{table:abltion_structure} shows the results on H3.6M dataset. "w/o $\mathcal{F}_{\rm OSTA}(\cdot)$" denotes not use the observed-only spatial-temporal attention. "Separate attention" denotes repeating $\mathcal{F}_{\rm OSTA}(\cdot)$ for $L$ times followed by $\mathcal{F}_{\rm FSTA}(\cdot)$ repeated for $L$ times. We can see that adding the observed-only spatial-temporal attention and using an iterative design on both attentions result in more effective motion prediction.



\begin{figure}[t]
\centering
\includegraphics[width=0.47\textwidth]{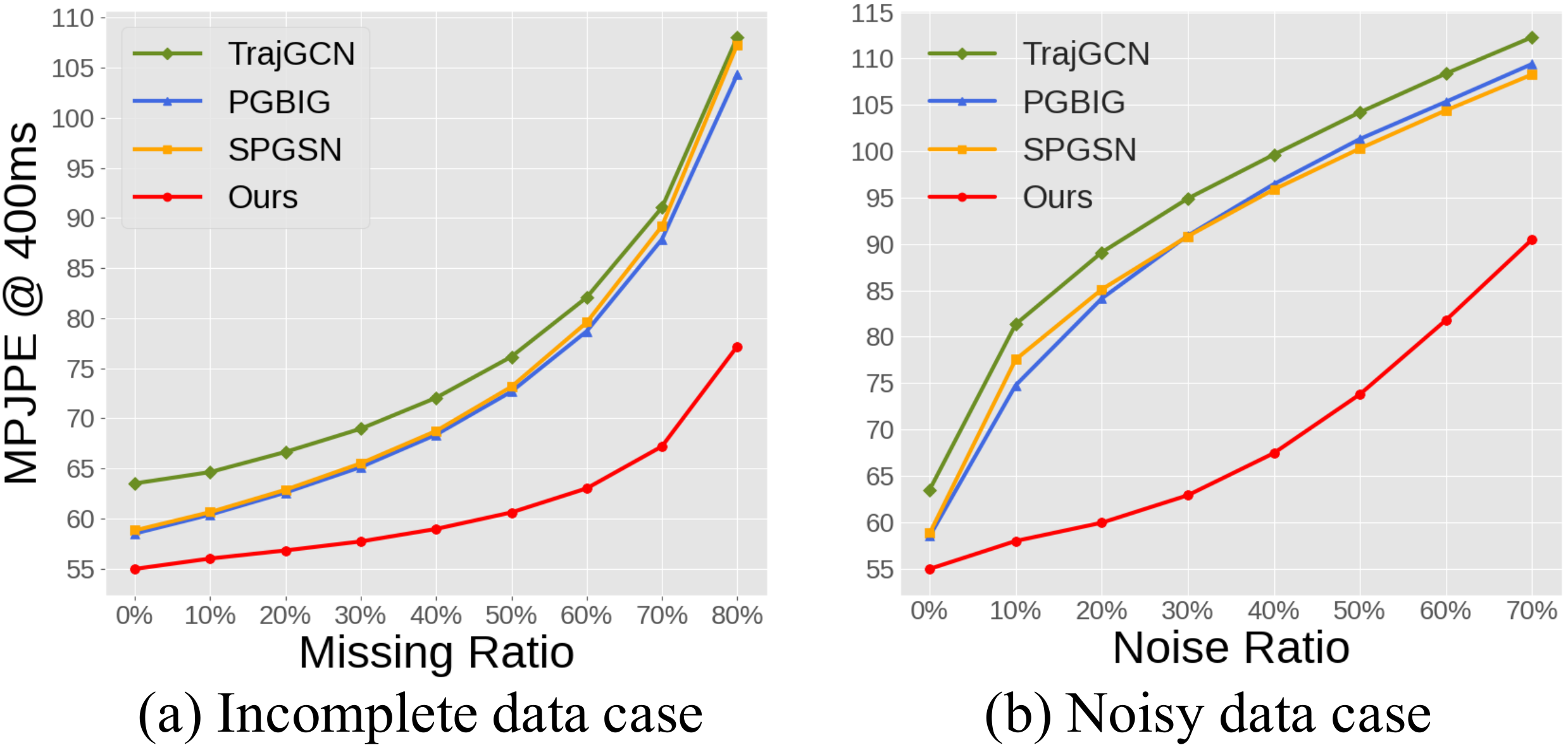}
\vspace{-3mm}
\caption{\small Comparison of model performance under missing data case and noisy data case on H3.6M.}
\label{fig:robust}
\vspace{-2mm}
\end{figure}

\vspace{0.1cm}
\noindent\textbf{Robustness analysis under incomplete \& noisy data cases.} We evaluate our method's robustness under two scenarios that frequently arise in real-world applications: incomplete and noisy data. For incomplete data, we randomly mask input motion coordinates during the testing phase, with various ratios.  To enable the baseline methods to handle missing values in the input motion, we apply linear interpolation and extrapolation to fill in the missing values before feeding the motion sequence into the model. Figure \ref{fig:robust} (a) shows the MPJPE at 400ms of our method and state-of-the-art baseline methods under different missing data ratios. We see that i) our method achieves the best prediction performance under all masking ratios; ii) with the increase of the masking ratio, the performance gap between our method and baseline methods widens, showing that our method is more robust with missing data than previous methods.

In the noisy data cases, we randomly add noises to different ratios of input motion coordinates in the test phase. Each coordinate's noise is sampled from Gaussian distribution $\mathcal{N}(0,50)$.
Figure \ref{fig:robust} (b) shows the comparison of MPJPE at 400ms under different noisy data ratios. We see that i) our proposed method achieves the best prediction performance under all noisy data ratios; ii) previous methods suffer greatly even when the noisy ratio is small, such as 10\% and 20\%, while our method's performance only experiences a slight drop in these cases and still maintains promising performance, showing better robustness.

\begin{table}[t]
  \centering
    \renewcommand\arraystretch{0.9}
    \setlength{\tabcolsep}{3pt}
  \footnotesize
  \caption{\small Ablation study on network structure design on H3.6M.}
    \vspace{-3mm}
  \small
    \begin{tabular}{l|cccccc}
    \toprule
        Ablation  & 80ms & 160ms & 320ms & 400ms & AVG \\
    \midrule
   w/o $\mathcal{F}_{\rm OSTA}(\cdot)$ & 9.7 & 21.6 & 46.0 & 57.3&  33.7 \\
    Separate attention &  9.5&20.7&44.1&55.1 & 32.4 \\
    Iterative attention (Ours) &\textbf{9.5} &\textbf{20.6}& \textbf{43.4} &\textbf{54.1} & \textbf{31.9}  \\
    \bottomrule
    \end{tabular}%
  \label{table:abltion_structure}%
  \vspace{-2mm}
\end{table}


\section{Conclusion}
In this work, we propose AuxFormer, a novel auxiliary model learning framework with an auxiliary-adaptive transformer network, to promote more comprehensive spatial-temporal dependency learning for 3D human motion prediction. The learning framework jointly learns the primary future prediction task with additional auxiliary tasks. To cooperate with the learning framework, we further propose an auxiliary-adapted transformer to model coordinate-wise spatial-temporal dependencies and be adaptive to missing data. We evaluate our method on three human motion prediction datasets and show our method achieves state-of-the-art prediction
performance and equips with higher robustness.

\section*{Acknowledgements}
This research is supported by NSFC under Grant 62171276, the Science and Technology Commission of Shanghai Municipal under Grant 21511100900 and 22DZ2229005,
and the National Research Foundation, Singapore under its Medium Sized Center for Advanced Robotics Technology Innovation. 
Robby T. Tan’s work is supported by MOE AcRF Tier 1, A-0009455-01-00.

{\small
\bibliographystyle{ieee_fullname}
\bibliography{egbib}
}

\end{document}